 \documentclass[final,3p,times,number, sort&compress]{elsarticle}

\usepackage{setspace}
\usepackage{algorithm}
\usepackage{algorithmicx}
\usepackage{algpseudocode}
\usepackage{amsmath}
\usepackage{amsthm}
\usepackage{framed}
\usepackage{xspace}
\usepackage{warpcol}

\usepackage{graphicx}
\usepackage{epsfig}
\usepackage{subfigure}
\usepackage{dcolumn}
\usepackage{float}
\usepackage[normalem]{ulem}
\useunder{\uline}{\ul}{}
\usepackage{multirow}
\usepackage{listings}
\usepackage{mathtools, amssymb}
\usepackage{cite}

\journal{Information Sciences}

\begin{document}

\begin{frontmatter}



\title{Global Context Aware RCNN for Object Detection}

\author[label1]{Wenchao Zhang}
\author{Chong Fu \fnref{label1,label2,label3} \corref{cor1}}
\ead{fuchong@mail.neu.edu.cn}
\cortext[cor1]{Corresponding author.}
\author[label1]{Haoyu Xie}
\author[label1]{Mai Zhu}
\author[label4]{Ming Tie}
\author[label5]{Junxin Chen}

\address[label1]{School of Computer Science and Engineering, Northeastern University, 
	Shenyang 110819, China}
\address[label2]{Engineering Research Center of Security Technology of Complex Network System, Ministry of Education, China}
\address[label3]{Key Laboratory of Intelligent Computing in Medical Image, Ministry of Education, Northeastern University, Shenyang 110819, China}
\address[label4]{Science and Technology on Space Physics Laboratory, Beijing 100076, China}
\address[label5]{College of Medicine and Biological Information Engineering, Northeastern University, Shenyang, 110819, China}

\begin{abstract}
RoIPool/RoIAlign is an indispensable process for the typical two-stage object detection algorithm, it is used to rescale the object proposal cropped from the feature pyramid to generate a fixed size feature map. However, these cropped feature maps of local receptive fields will heavily lose global context information. To tackle this problem, we propose a novel end-to-end trainable framework, called Global Context Aware (GCA) RCNN, aiming at assisting the neural network in strengthening the spatial correlation between the background and the foreground by fusing global context information. The core component of our GCA framework is a context aware mechanism, in which both global feature pyramid and attention strategies are used for feature extraction and feature refinement, respectively. Specifically, we leverage the dense connection to improve the information flow of the global context at different stages in the top-down process of FPN, and further use the attention mechanism to refine the global context at each level in the feature pyramid. In the end, we also present a lightweight version of our method, which only slightly increases model complexity and computational burden. Experimental results on COCO benchmark dataset demonstrate the significant advantages of our approach.

\end{abstract}




\end{frontmatter}


\section{Introduction}
\begin{figure*}[htbp]
  \centering
    \subfigbottomskip=0.1pt{}
    \subfigure{\includegraphics[width=0.22\textwidth]{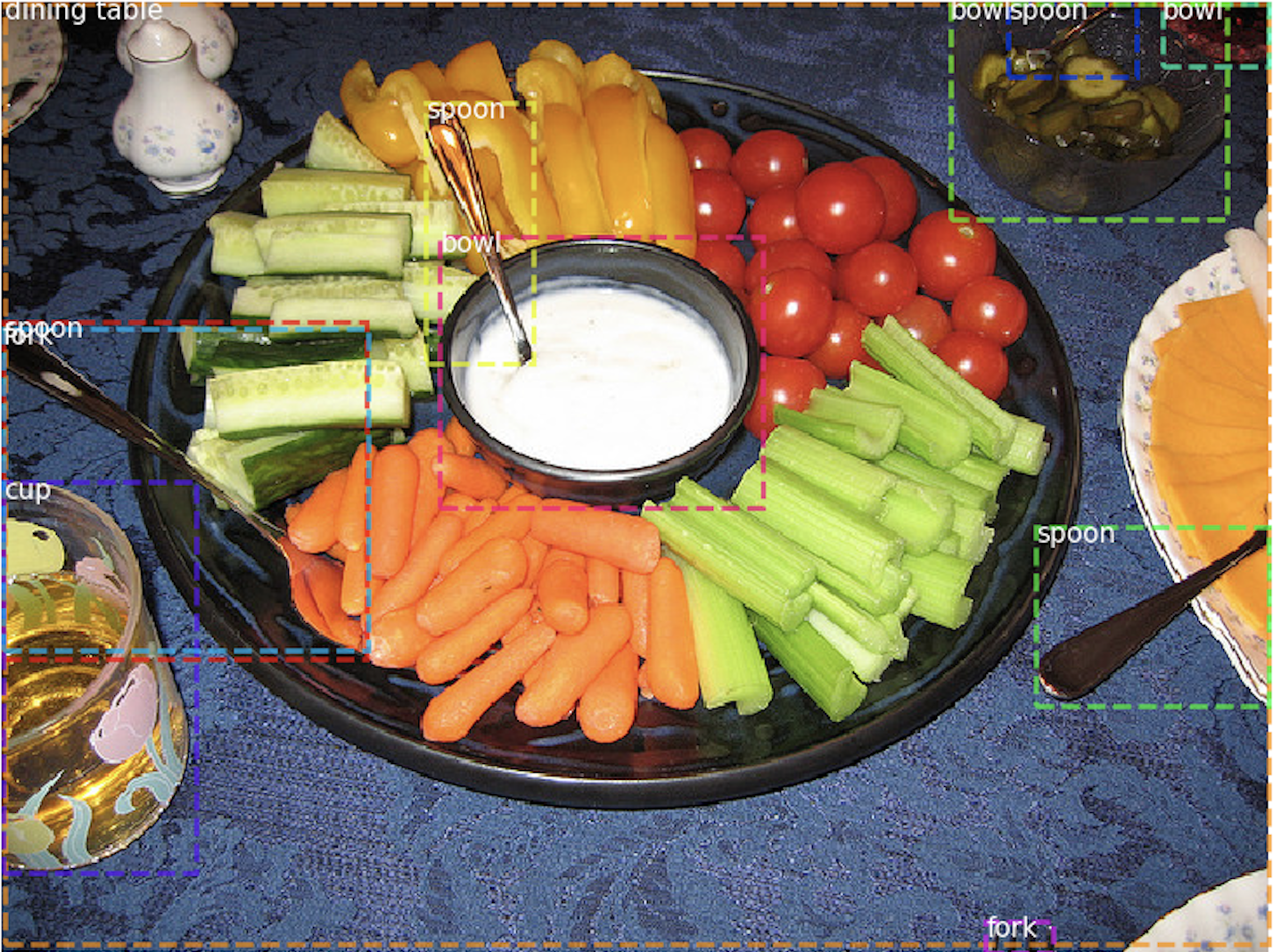}}  
  \subfigure{\includegraphics[width=0.22\textwidth]{./figures/table5.png}} 
    \subfigure{\includegraphics[width=0.22\textwidth]{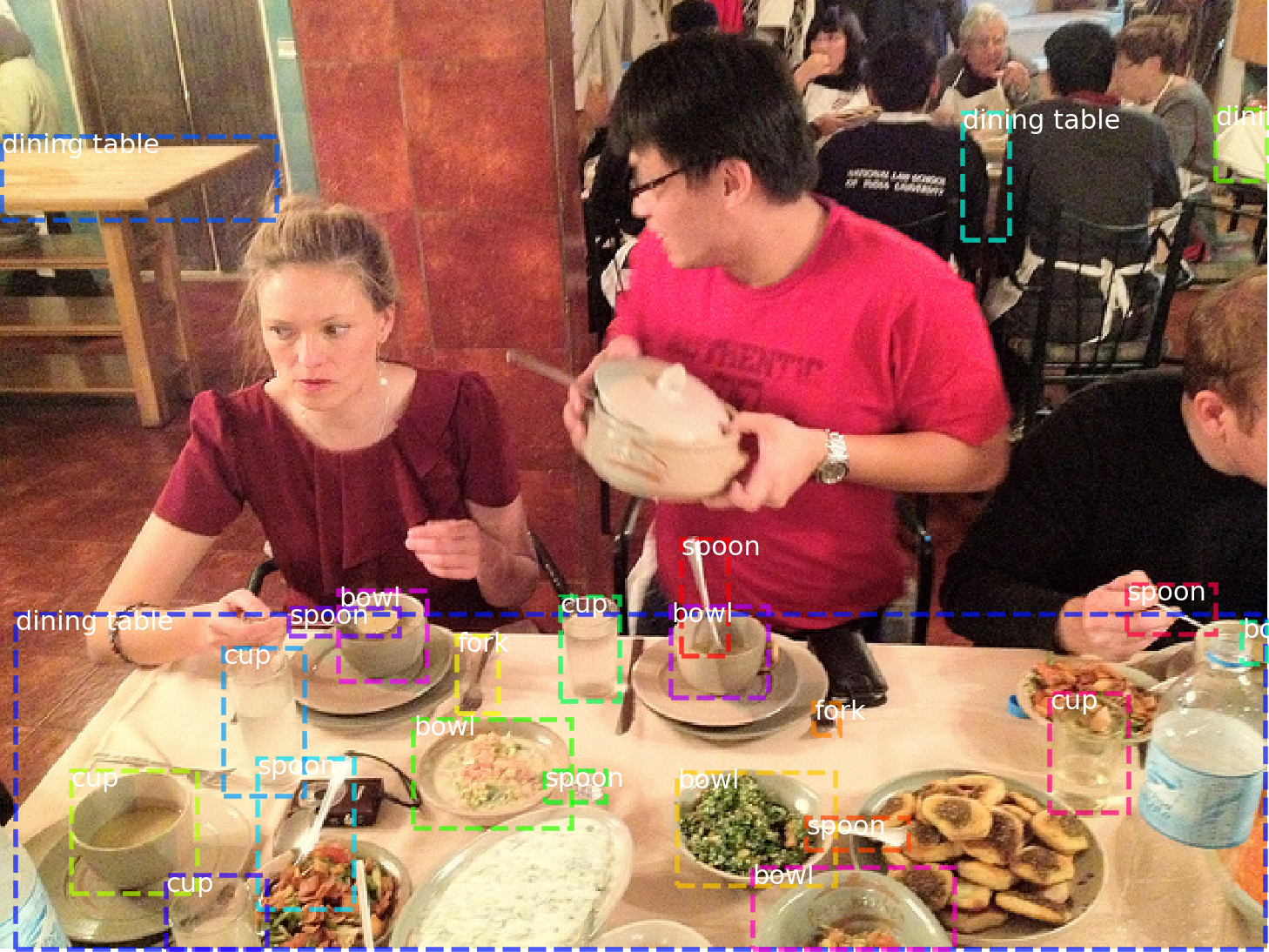}} 
    \subfigure{\includegraphics[width=0.22\textwidth]{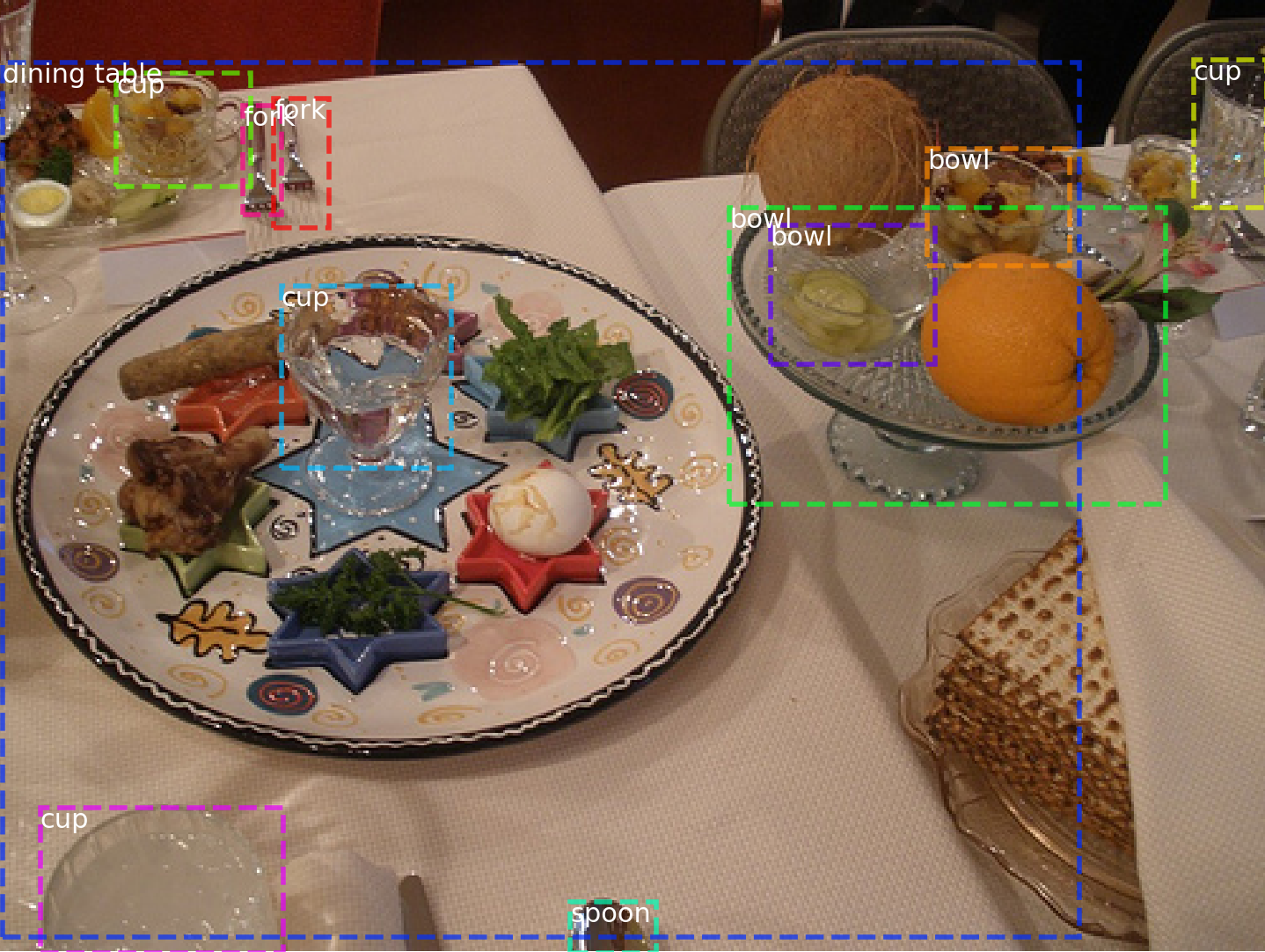}} \\
  \subfigure{\includegraphics[width=0.22\textwidth]{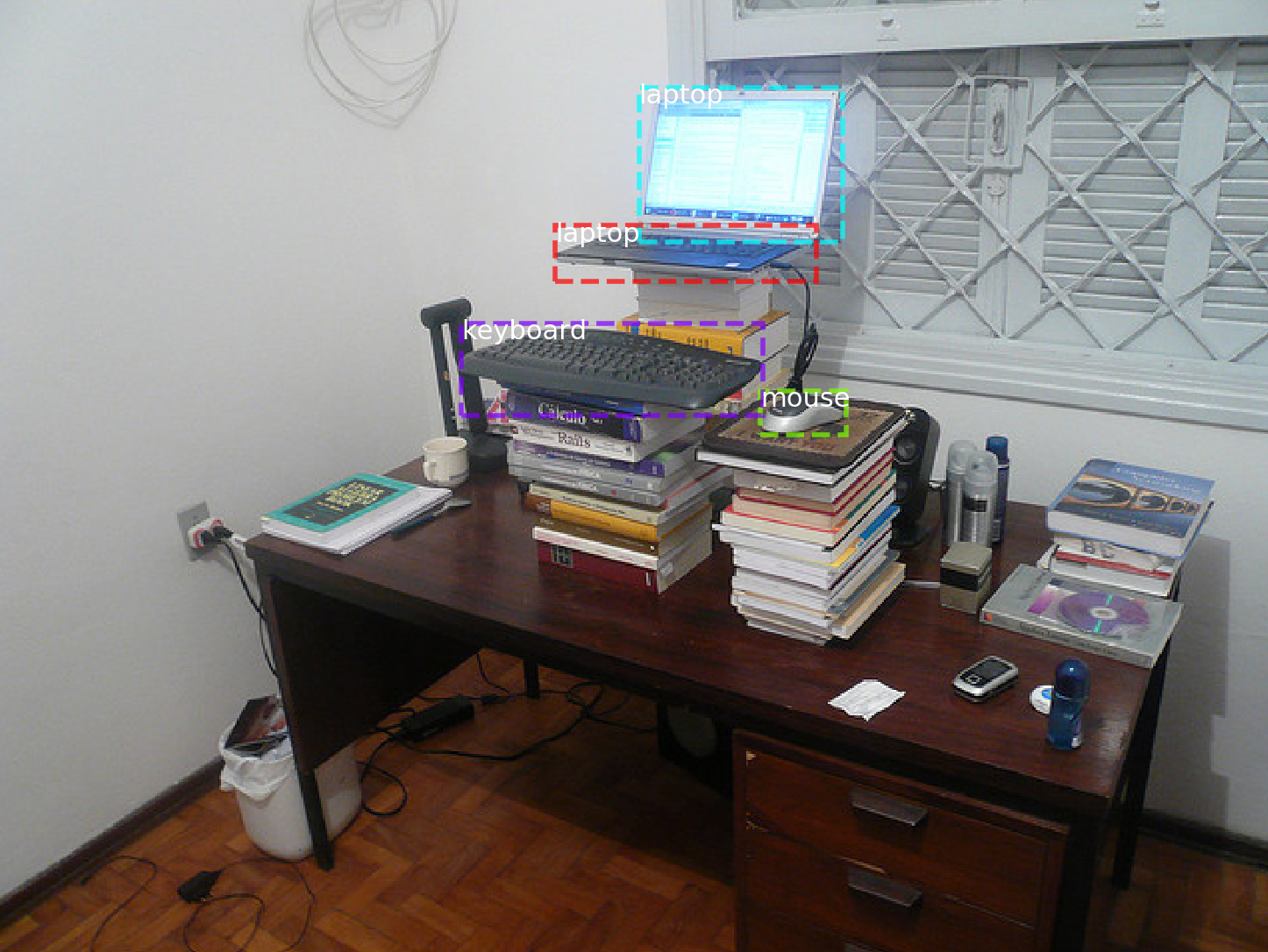}} 
    \subfigure{\includegraphics[width=0.22\textwidth]{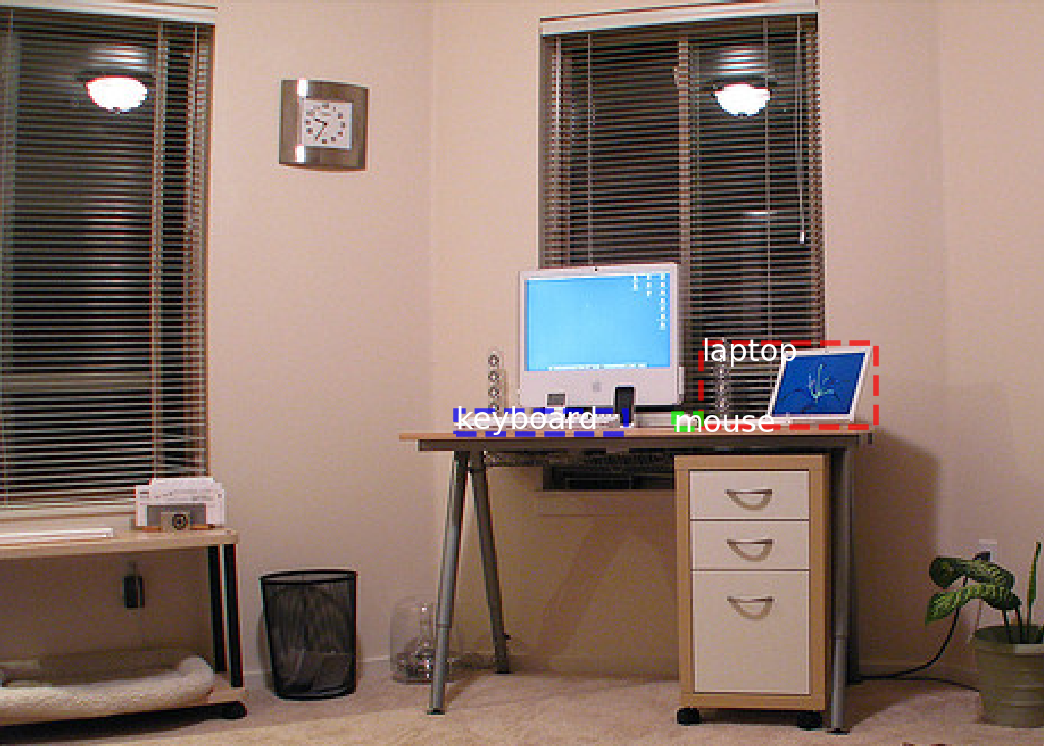}} 
    \subfigure{\includegraphics[width=0.22\textwidth]{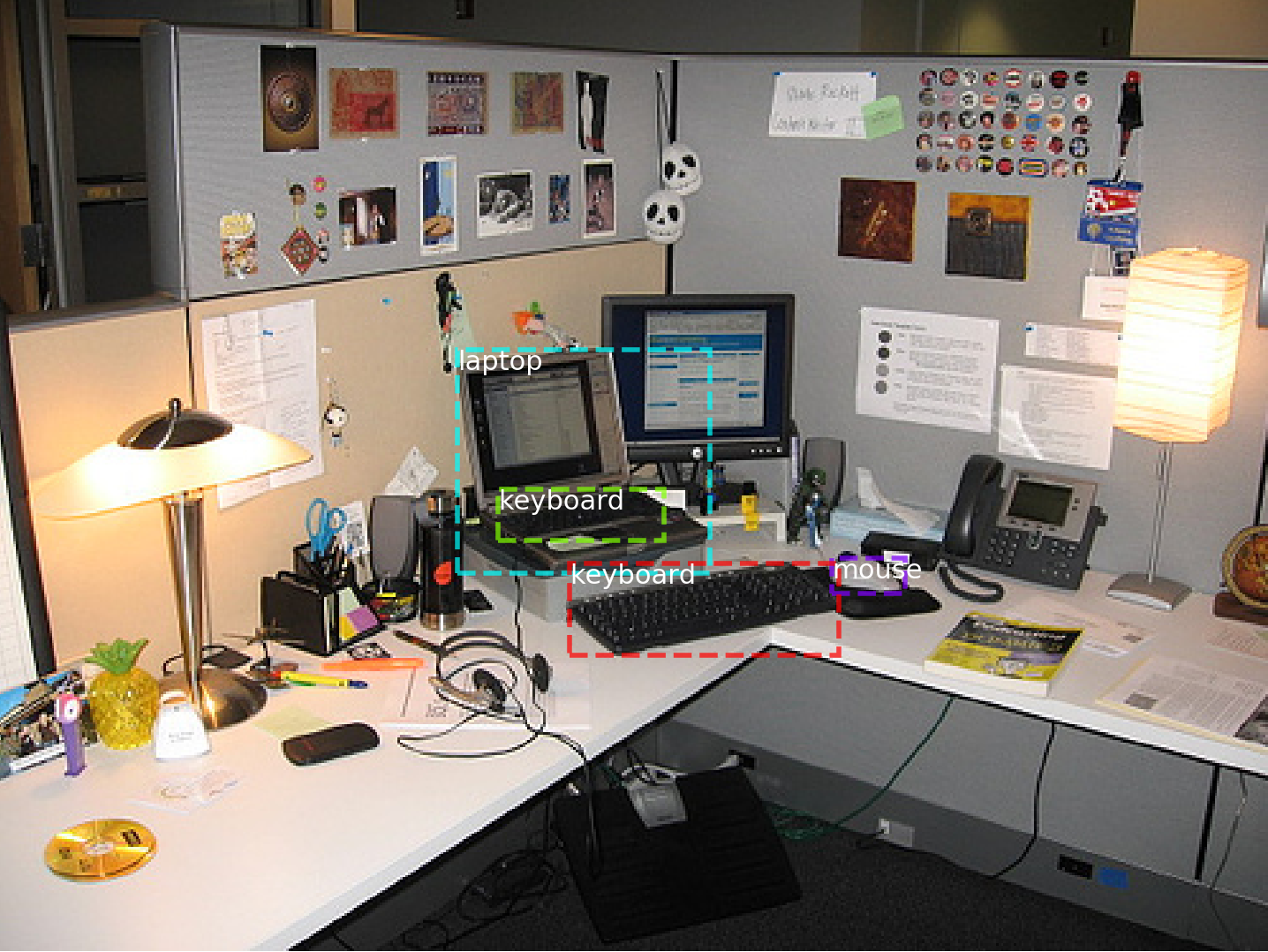}}  
  \subfigure{\includegraphics[width=0.22\textwidth]{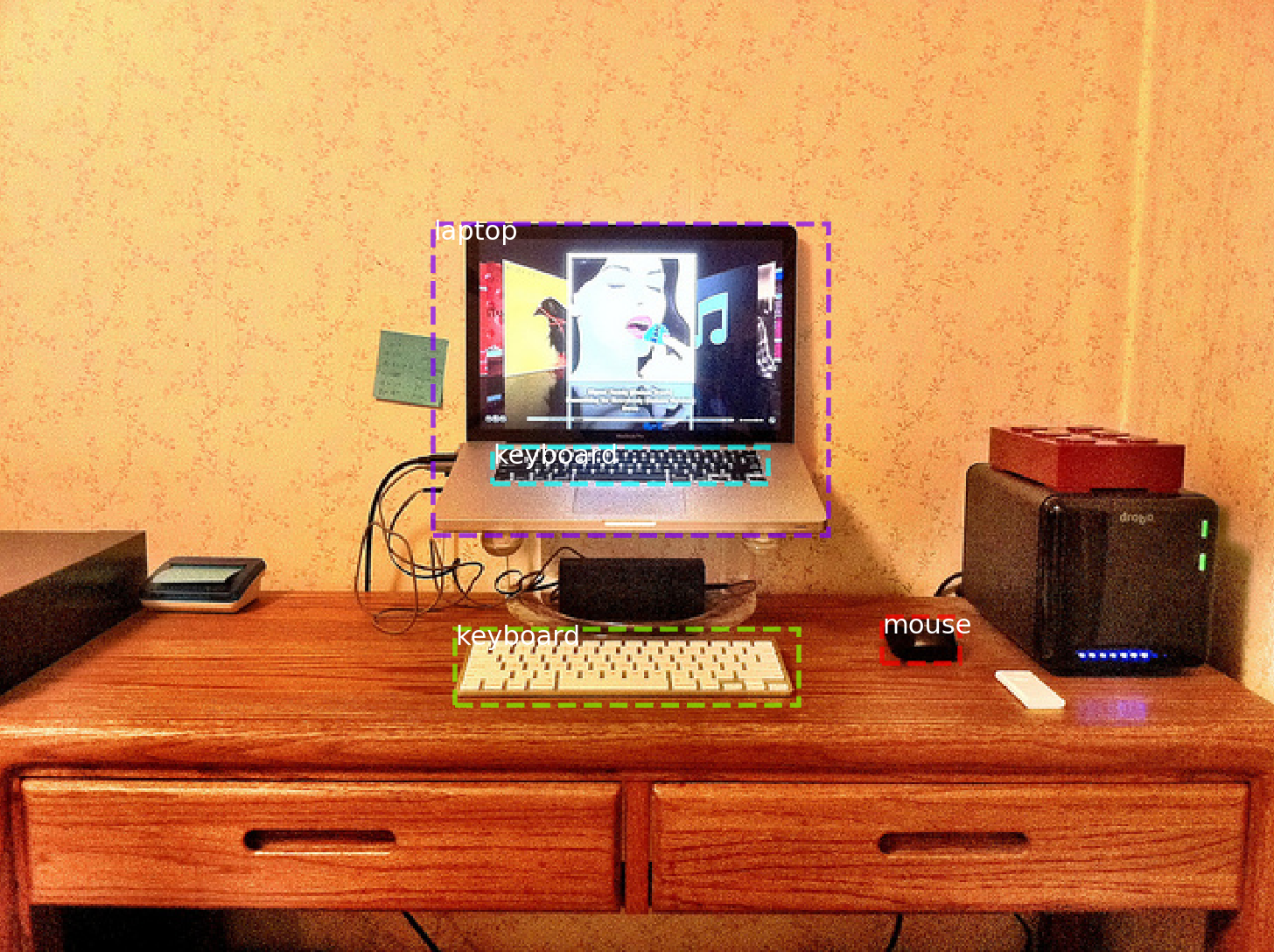}} \\
    \caption{Examples of potential relationship between global context and local information, top row: some kitchen utensils such as knives, forks, bowls and cups often appear on the table with food; bottom row: computer, keyboard and mouse often appear together.}
  \label{fig:fig1}
\end{figure*}

Benefiting from the development and application of deep network technology in computer vision community, the performance of wide range of computer vision tasks such as target detection, semantic segmentation and instance segmentation have been greatly improved. In recent years, many excellent detection frameworks have been proposed. For example, there are one-stage methods with faster speed such as SSD~\citep{liu2016ssd} and YOLO~\citep{redmon2016you}, and two-stage methods with better detection performance such as Faster RCNN~\citep{ren2015faster} and FPN~\citep{lin2017feature}. It is a remarkable fact that most of the currently popular two-stage methods usually use RoIPool/RoIAlign to align regions of interest of different scales to meet the requirement of consistent input size of the neural network. In FPN, it adaptively crops the regions of interest from the feature pyramid corresponding to different spatial scales, and then uniformly resizes them to a fixed spatial scale of $7\times7$ through RoIPool/RoIAlign, and after flattening these feature maps, they are further encoded through two fully connected layers, and finally classification and positioning tasks are performed respectively. However, is it reasonable to use only local receptive field features such as object proposals for classification and positioning? On the one hand, for classification tasks, the target object has a natural inter-dependencies relationship with the background and other foreground objects. For example, in real scenario, cups often appear on the dining table, and there are food, knives, forks and bowls around them, in addition laptop, keyboard, mouse often appear together on the desk, as shown in the Fig.~\ref{fig:fig1}. On the other hand, for positioning tasks, the candidate box coordinates predicted by the detection model are the relative positions in the whole image, so some references in the background can help to locate the target. Therefore, simply using the feature maps of object proposals for detection will bring about the loss of the spatial and category relationship information between these local and global contexts. 

To mitigate the drawback mentioned above, in this paper, we propose a context aware mechanism that allows the two-stage object decetion network to fuse the global context information with the local informations of the RoIs (Regions of Interest). In two-stage methods, rpn head (i.e., the first stage) and roi head (i.e., the second stage) both use the image features extracted by the backbone network for prediction tasks, where rpn head is responsible for distinguishing the foreground and background, and predicting the regression coefficient of the anchor box, while roi head is responsible for predicting the specific category of roi and obtaining the offset value used to fine-tune the bounding box. Therefore, we enhance its global feature perception capabilities in these two stages respectively. For rpn head, it uses the overall features of the output from the backbone network for prediction tasks, so we only use global feature statistics to perform feature calibration. But for roi head, it uses partial features cropped from the overall feature map as input, so we need a more sophisticated design for the extraction of its global features. The structure of the our model is depticted in Fig.~\ref{fig:fig3}. Specifically, we believe that the feature maps at different stages in the feature pyramid carry global context information of different attributes. Therefore, in order to make full use of this information to help the neural network better complete the object detection task, we fuse the global context information of different stages through dense connection in the roi head, and then leverage our proposed context aware module to generate higher-dimensional global descriptors. Simultaneously, like FPN, before decoupling the positioning and classification tasks, we use two shared fully connected layers to further extract features at different stages, and finally we fuse the prediction information of different stages to make the final decision.

To provide evidence for these claims, in Section \ref{expe} we develop several ablation studies and conduct an extensive evaluation on the COCO dataset~\citep{lin2014microsoft}. We also present results beyond COCO that indicate that the benefits of our approach are not restricted to a specific dataset. In the end, our method gains +1.5 and +0.9 AP on MS COCO dataset from Feature Pyramid Network (FPN) baselines with ResNet-50 and ResNet-101 backbones, respectively.

In summary, the main contributions of this work are highlighted as follows:

  1. We observe that due to the use of RoIPool/RoIAlign, a lot of global context information will be lost in most two-stage object decetion network, hence we propose GCA RCNN to enhance and refine global context information by using dense connection and attention mechanism, and finally the missing global information is compensated by the fusion of global context and local features.

  2. Unlike SENet~\citep{hu2018squeeze}, which uses global context information for feature recalibration at the convolutional level, we extend it to assemble global context information on the two-stage target detection pipeline. 
  
  3. Our method can be easily deployed in other FPN based methods, and we also present a lightweight version of our method, which has a small amount of additional computing overhead.

The rest of this paper is organized as follows. In Section \ref{related}, we briefly review related work on object detection and context awareness. In Section \ref{method}, we introduced our method in detail from dense global context, context awareness, and feature fusion. Experimental details and analysis of the results are elaborated in Section \ref{expe}. Finally, we conclude the paper in Section \ref{conclusion}.

\section{Related Work}
\label{related}


\subsection{Object Detection}
There are two common ways for object detection: one-stage and two-stage. Classic one-stage methods such as SSD~\citep{liu2016ssd},YOLO~\citep{redmon2016you,redmon2017yolo9000,redmon2018}, etc. quickly classify and locate targets in an end-to-end manner. Classic two-stage methods such as Faster RCNN~\citep{ren2015faster}, FPN~\citep{lin2017feature}, etc. first obtain object proposals through the RPN (Region Proposal Network), and then use RoIPool/RoIAlign to align the spatial scales of these object proposals before performing detection. Most of the subsequent algorithms are also based on these two structures for continuous improvement and development. CornerNet~\citep{law2018cornernet}, ExtremeNet~\citep{zhou2019bottom}, CenterNet~\citep{zhou2019objects}, and FCOS~\citep{tian2019fcos}, etc., use keypoint detection technology to optimize the anchor generation process. Wu $et~al$.~\citep{wu2020rethinking} study that convolutional neural networks and fully connected networks have different sensitivity to classification tasks and positioning tasks, therefore, it decouples the positioning and classification tasks of FPN to improve detection performance. Wang $et~al$.~\citep{wang2019region} proposed to guide the generation of anchor through image features. Cascade r-cnn~\citep{cai2018cascade} considers that training samples under different IoU (Intersection over Union) conditions have different effects on the performance of target detection network, and then proposes the structure of muti stage, in which each stage has a different IoU threshold. HTC~\citep{chen2019hybrid} tries to integrate semantic segmentation into the instance segmentation framework to obtain a better spatial context. DetectoRS~\citep{qiao2020detectors} proposed RFP (Recursive Feature Pyramid) and SAC (Switchable Atrous Convolution) to realize looking and thinking twice or more. Although the two-stage methods mentioned above can achieve excellent detection performance, they inevitably use RoIPool/RoIAlign in roi head, resulting in the lack of global context.

\subsection{Context Awareness}
Assembling global context information of the target can enable the neural network to learn more about the relationship between the foreground and the background, so that it can rely on this potential relationship feature to help the neural network highlight and identify the target. There are many ways to obtain contextual information, an alternative way is to use the attention mechanism to obtain global context information~\citep{bello2019attention,vaswani2017attention,woo2018cbam,hu2018squeeze,hu2018relation} and use it for feature recalibration. Alongside the methods described above, there is also another way to use contextual information, for the two stage object detection method, the target needs to be aligned to a uniform scale through RoIPool/RoIAlign, the general way is to expand the object proposal by a few pixels when cropping the target from the feature map to obtain more surrounding information~\citep{tang2018pyramidbox,wu2020context}. Context information can also be used in many other ways. Lin $et~al$.~\citep{lin2019context} proposed CGC (Context-Gated Convolution) to adaptively modify the weight of the convolutional layer. Si $et~al$.~\citep{si2018dual} proposed DuATM (Dual Attention Matching network) to learn context-aware feature sequences, and perform pedestrian re-identification by performing sequence comparison simultaneously. Although the above-mentioned methods verify the importance of the global context, they do not involve the relationship between the cropped local features obtained after RoIPool/RoIAlign and their corresponding global contexts. 

\begin{figure}[htbp]
\centering
\includegraphics[width=0.6\textwidth]{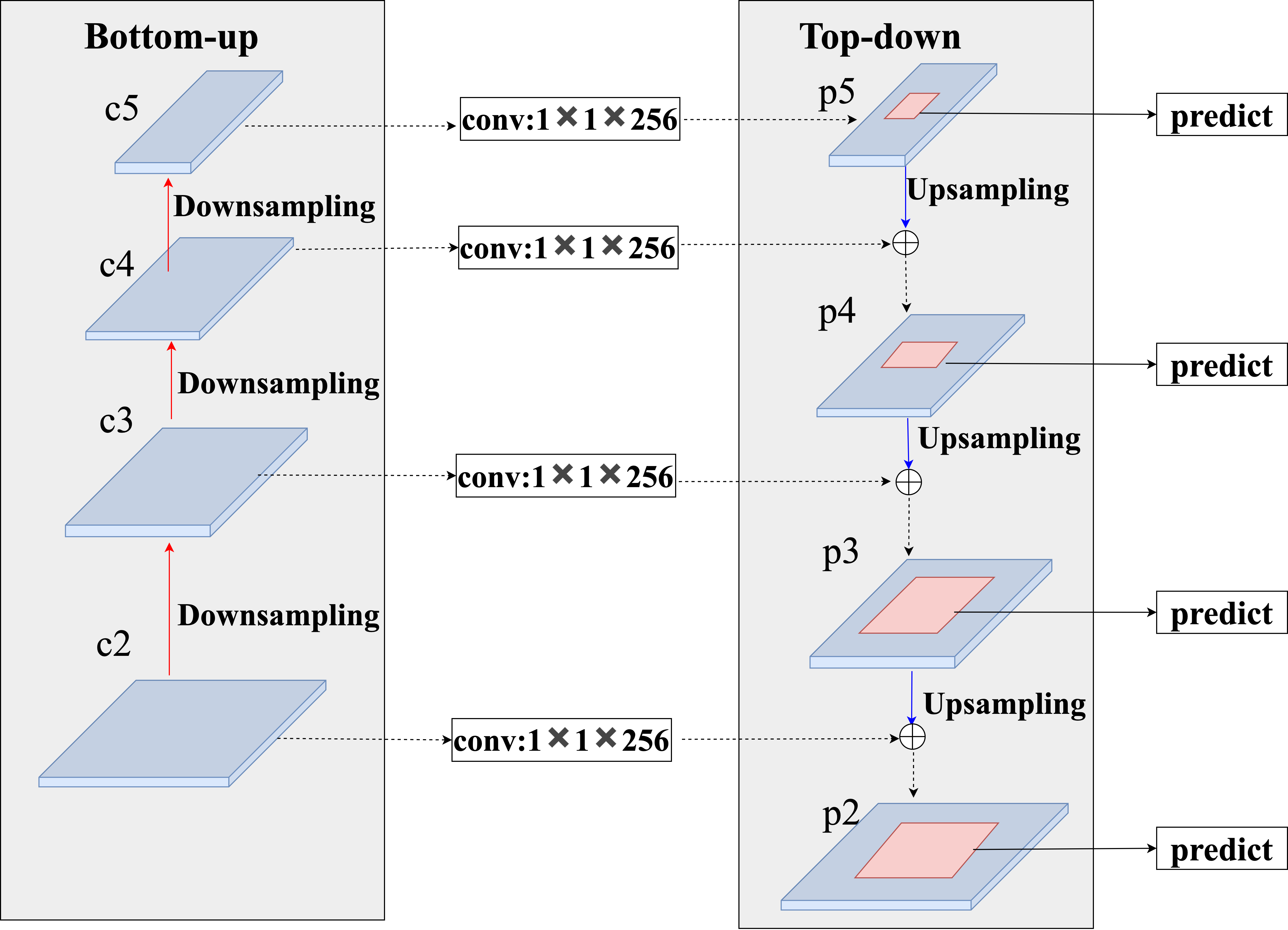}

   \caption{Network architecture of Feature Pyramid Network (FPN). In this Figure, the upward red arrow represents the down-sampling process, the downward blue arrow represents the up-sampling process, the black dashed line represents the data flow, and $\oplus$ represents the element-wise sum.}
\label{fig:short}
\label{fig:fig2}
\end{figure}

\section{Method}
\label{method}
 In this Section, we will first introduce the motivation of our design, and then describe the structure of our model in detail from three aspects: dense global context, context awareness, and feature fusion. Note that since in rpn head does not involve RoIAlign, we will introduce the enhancement of global context information for rpn head in Section \ref{expe}.
\subsection{Motivation}
 In FPN, the feature pyramid is constructed through Bottom-up pathway, Top-down pathway and lateral connections. As shown in Fig.~\ref{fig:fig2}, where Bottom-up pathway refers to the process of down-sampling the input image 5 times in the backbone network, and the output of the residual blocks corresponding to $\displaystyle \{conv2, conv3, conv4, conv5\}$ is denoted as $\displaystyle \{c2, c3, c4, c5\}$, where Top-down pathway refers to the up-sampling process after convoluting c5 by $1\times1$ convolutional layer, for simplicity, we denote the final feature map set as $\displaystyle \{p2, p3, p4, p5\}$, where lateral connection refers to the process of fusing the corresponding feature maps between $\displaystyle \{c2, c3, c4, c5\}$ and $\displaystyle \{p2, p3, p4, p5\}$ through the $1\times1$ convolutional layer. In the roi head of FPN, only using the cropped feature map of object proposal to locate and classify the object will greatly lose the global context information, which will lead to the loss of the convolutional neural network's ability to perceive the relationship between the background and the foreground information. Therefore, we design our method from two aspects: the acquisition of global context information and the fusion of local and global information.

\subsection{Dense global context}
 
In order to unify the spatial scale of the global context information at different stages in the feature pyramid, we leverage adaptive average pooling to downsample the spatial scale of $\displaystyle \{p2, p3, p4, p5\}$ to $(M,N)\times \displaystyle \{1, 1/2, 1/4, 1/8\}$ respectively. Then we continue to downsample these pooled feature maps using four parallel branches which containing $\displaystyle \{3, 2, 1, 0\}$ downsampling blocks respectively, each downsampling block refers to the composite function of two consecutive operations: a $3\times3$ convolution (conv) with stride 2 followed by a ReLU (rectified linear unit)~\citep{nair2010rectified} activation function, for simplicity, we denote the downsampling block as \emph{D}. To this end, the global context feature map set obtained by dense connection is denoted as $\displaystyle \{g0, g1, g2, g3\}$. As a consequence, the benefits of the global context captured by multi-branch downsampling blocks can be accumulated through the network. The output feature $g_i$ is defined by 
\begin{displaymath}{}
g_i=D^{3-i}([\phi(p_{i+2}), W_i]),\tag{$1$}{}
\end{displaymath}

\begin{displaymath}
W = \begin{pmatrix} 0 & 0 & 0 \\ 1 & 0 & 0 \\ 1 & 1 & 0 \\ 0 & 0 & 1 \end{pmatrix}\begin{pmatrix} D^i(\phi(p_{2})) & D^{1}[\phi(p_{3}),D^{1}(\phi(p_{2}))] & [g_0,g_1,g_2]  \end{pmatrix}^{T},\tag{$2$}
\end{displaymath}
where $D^i$ represents the total use of i downsampling blocks, $\phi$ stands for adaptive average pooling function, and where $W_i$ represents the i-th row of matrix $W$. After getting the feature set $\displaystyle \{g0, g1, g2, g3\}$ of the global context information, we input $g_i$ and the feature maps of object proposal respectively into four parallel context aware modules. Motivated by ~\citep{huang2017densely}, we define $[\phi(p_i),W_i]$ as a concatenation of the feature-maps in it.

\begin{figure*}
\begin{center}
\includegraphics[width=15cm,height=7.5cm]{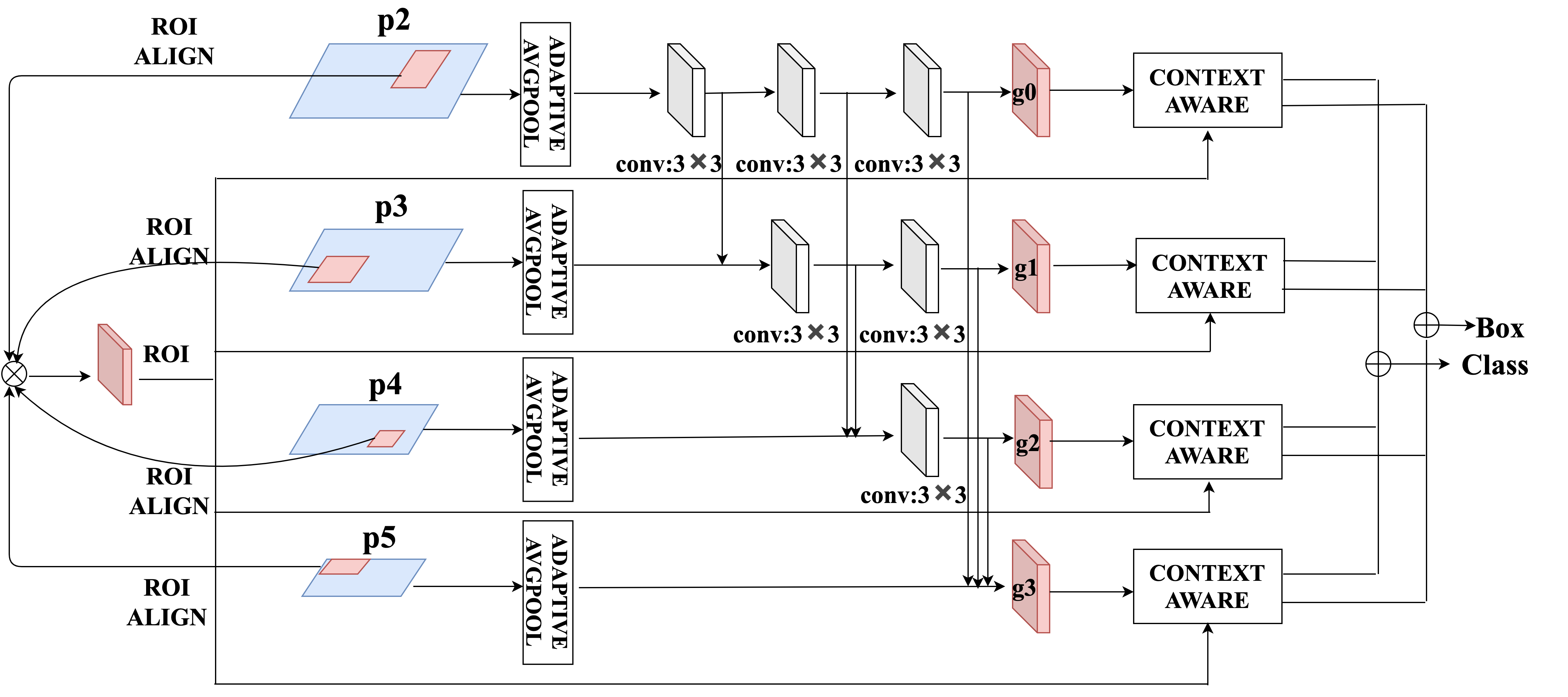}
\end{center}
   \caption{Network architecture of GCA RCNN (We only show the architecture of roi head on the figure for neat presentation). In this Figure, $\displaystyle \{p2, p3, p4, p5\}$ represents the feature pyramid used to generate RoI from the corresponding feature map and the channel dimensions are all 256, $\displaystyle \{g0, g1, g2, g3\}$ represents the global context feature map obtained by down-sampling $\displaystyle \{p2, p3, p4, p5\}$ through the corresponding $3\times3$ convolutional layer, $\otimes$ represents the concatenation function and $\oplus$ represents the element-wise sum.}
\label{fig:short}
\label{fig:fig3}
\end{figure*} 

\subsection{Context awareness}

\begin{figure*}
\begin{center}
\includegraphics[width=14cm,height=6cm]{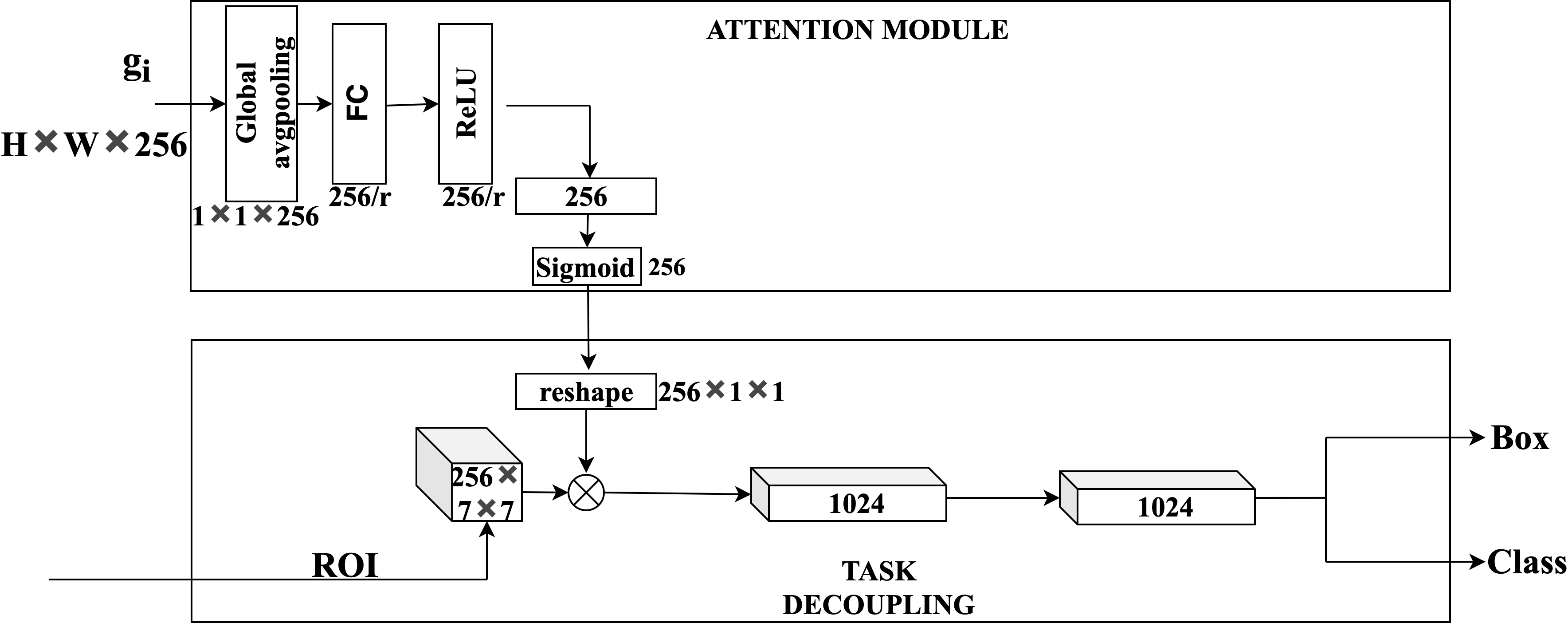}
\end{center}
   \caption{Network architecture of context aware module. It contains two sub-modules: attention and task decoupling, the attention module is in charge of mappings global context information to high-dimensional space and the task decoupling module outputs the prediction results of classification and positioning. In this Figure, $\otimes$ represents the channel-wise multiplication.}
\label{fig:short}
\label{fig:fig4}
\end{figure*}

A diagram illustrating the structure of an context aware module is shown in Fig.~\ref{fig:fig4}, the context aware module consists of two sub-modules: attention module and task decoupling module. In the attention module, inspired by ~\citep{hu2018squeeze}, we embedding these global context information $\displaystyle \{g0, g1, g2, g3\}$ into higher-dimensional features for characterizing the global features. Specifically, we leverage the global average pooling layer to squeeze the spatial scale of $\displaystyle \{g0, g1, g2, g3\}$ that produces a channel-wise statistics by aggregating feature maps across their spatial dimensions, then we leverage a bottleneck block composed of two fully connected layers with reduction of r to squeeze-and-excitation the dimension of the feature map and learn a non-mutually-exclusive relationship between global context information and local information, note that in the bottleneck block, the first fully connected layer uses the ReLU activation function, and the second fully connected layer uses the sigmoid activation function. In the task decoupling module, we first do the channel-wise multiplication between the global context descriptor obtained by the squeeze-excitation module and the $256\times7\times7$ object proposal feature tensor after RoIAlign resize, and then in order to further combine features of different attributes and enhance generalization, after flattening the object proposal feature map, like FPN, we used two fully connected layers with an output dimension of 1024. It is worth noting that, different from SENet multiplying the global descriptor on each channel of the squeeze-and-excitation block input feature map to perform feature recalibration, we multiply it on each channel of the $256\times7\times7$ feature tensor obtained by RoIAlign to strengthen the mutual relationship between global context and local receptive field.

\subsection{Feature fusion}

Different from the box head in FPN, the object proposal rescaled by RoIAlign will be classified and located after the subsequent two fully connected layers. In our approach, we leverage four parallel branches to extract global context information at different stages in the feature pyramid, and further leverage these global context informations to fusion with the local receptive fields of object proposals in each branch through the attention mechanism. Simultaneously, like FPN, after each branch, we decouple these fused feature information through two parallel fully connected layers for the classification task and the positioning task. Finally, the two parallel 1D features output by each branches are fused through the element-wise sum for the final prediction.




\section{Experiments}
\label{expe}

In this Section, we will first introduce our experimental dataset, evaluation criteria, and model parameter settings. Then we conducte ablation experiments on COCO dataset from two aspects, namely the dense connection and global context awareness. Moreover, we verified the generalization of our method on the Cityscapes dataset.

\subsection{Dataset and metrics}
We verify our approch on the large scale detection benchmark COCO dataset with 80 object categories, which are splited into 115k, 5k and 41k images for train/minival/test. Because the labels of its test-dev split are not publicly available, we use its minival dataset for our ablation study. Simultaneously, we present our final results on the test-dev split (20K images) by uploading our detection results to the evaluation server. We use the COCO standard metric to evaluate the AP under different IoU [0.5:0.05:0.95], and finally take the average of APs under these thresholds as the result, denoted as mAP@[.5, .95 ].

\subsection{Implementation Details}
Our model is end-to-end trained based on the torchvision detection module~\citep{paszke2019pytorch}, using SGD with 0.9 momentum, 0.0005 weight decay for gradient optimization. We train detectors on a single NVIDIA titan xp GPU with the mini-batch size of one image. Unless specified, ResNet-50 pretrained on imagenet~\citep{imagenetcvpr09} is taken as the backbone networks on this dataset. Following the common practice, the size of the input image is adjusted to 800 for the short side and less or equal to 1333 for the long side. We train detectors for 24 epochs with an initial learning rate of 0.0025, and decrease it by 0.1 after 16 and 22 epoches, respectively. For data augmentation, we randomly flip the input image horizontally with a probability of 0.5. And all newly added convolutional layers are randomly initialized with the “xavier” method~\citep{glorot2010understanding}.



\subsection{Ablation study}

{\bf Dense connection:}
Table~\ref{tab:my-table1} shows the impact of whether the global context features $\displaystyle \{p2, p3, p4, p5\}$  at different stages in the feature pyramid are densely connected on the performance of our proposed model, it should be noted that we did not use the attention mechanism in this section and the initial adaptive pooling size is (64, 96). Specifically, in order to further encode local receptive field information and global context information, after the 1D global context information captured through global average pooling in attention module and the object proposal $256\times7\times7$ feature map tensor in task decoupling module, we connect a fully connected layer with an output dimension of 512 respectively. Finally, we concatenate the two pieces of 512 1D tensor together, and then connect a fully connected layer with an output dimension of 1024 to learn the potential relationship between local receptive fields and global context information. From Table~\ref{tab:my-table1} we can infer that the use of dense connections can enhance the global information flow between different stages, so the model performance is better improved (outperforms FPN basline by 1.0\% on COCO's standard AP metric and by 2.0\% on AP@IoU=0.5). We used dense connections in all subsequent experiments.
\begin{table}[t]

\caption{Effect of dense connection on COCO val2017 (\%)}
\label{tab:my-table1}
\setlength{\tabcolsep}{2mm}{
\begin{center}
\begin{tabular}{cccc}
\hline
\multicolumn{1}{c}{\bf Method}  &\multicolumn{1}{c}{\bf AP} &\multicolumn{1}{c}{\bf AP$_{0.5}$} &\multicolumn{1}{c}{\bf AP$_{0.75}$}  
\\ \hline 
FPN baseline     & 36.8 & 58.0  & 40.0   \\
Dense connection (ours) &  \textbf{37.8} &  \textbf{60.0}  &  \textbf{40.6}   \\ 
\hline
\end{tabular}

\end{center}}

\end{table}

{\bf Dense connection with different pooling size: }
Table~\ref{tab:my-table2} shows the impact of dense connection of global context information at different stages in the feature pyramid on the performance of the model under different pooling size conditions. Assuming that our initial adaptive pooling size is (128, 192), thus the pooling sizes corresponding to the four stages $\displaystyle \{p2, p3, p4, p5\}$ from the bottom to the top of the FPN are $(128, 192)\times \displaystyle \{1/8, 1/4, 1/2, 1\}$, with the goal of balancing the memory consumption and accuracy, we analyze the model performance when the initial pooling size are $(128, 192)\times \displaystyle \{1, 1/2, 1/4, 1/8\}$ respectively. The comparison in Table~\ref{tab:my-table2} shows that performance dose not continuously increasing with bigger pooling size, we find that setting the initial pool size to (64, 96) achieve a good balance between memory consumption and accuracy, hereafter we used this value in all experiments.

\begin{table}[t]
\centering
\caption{Effect of dense connection with different pooling size on COCO val2017 (\%)}
\label{tab:my-table2}
\setlength{\tabcolsep}{4mm}{
\begin{tabular}{cccc}
\hline
\multicolumn{1}{c}{\bf Pooling Size}  &\multicolumn{1}{c}{\bf AP} &\multicolumn{1}{c}{\bf AP$_{0.5}$} &\multicolumn{1}{c}{\bf AP$_{0.75}$} 
\\ \hline 
(128,192)    & 37.8 & 60.0  & 40.6   \\
(64,96)      &  \textbf{38.0} &  \textbf{60.1}  &  \textbf{41.0}   \\
(32,48)      & 37.8 & 59.7  & 40.9   \\
(16,24)      & 37.7 & 59.7  & 40.2   \\ 
\hline
\end{tabular}}

\end{table}

\begin{figure*}[!htb]
  \centering
    \subfigure[]{\includegraphics[width=6.3cm,height=3.7cm]{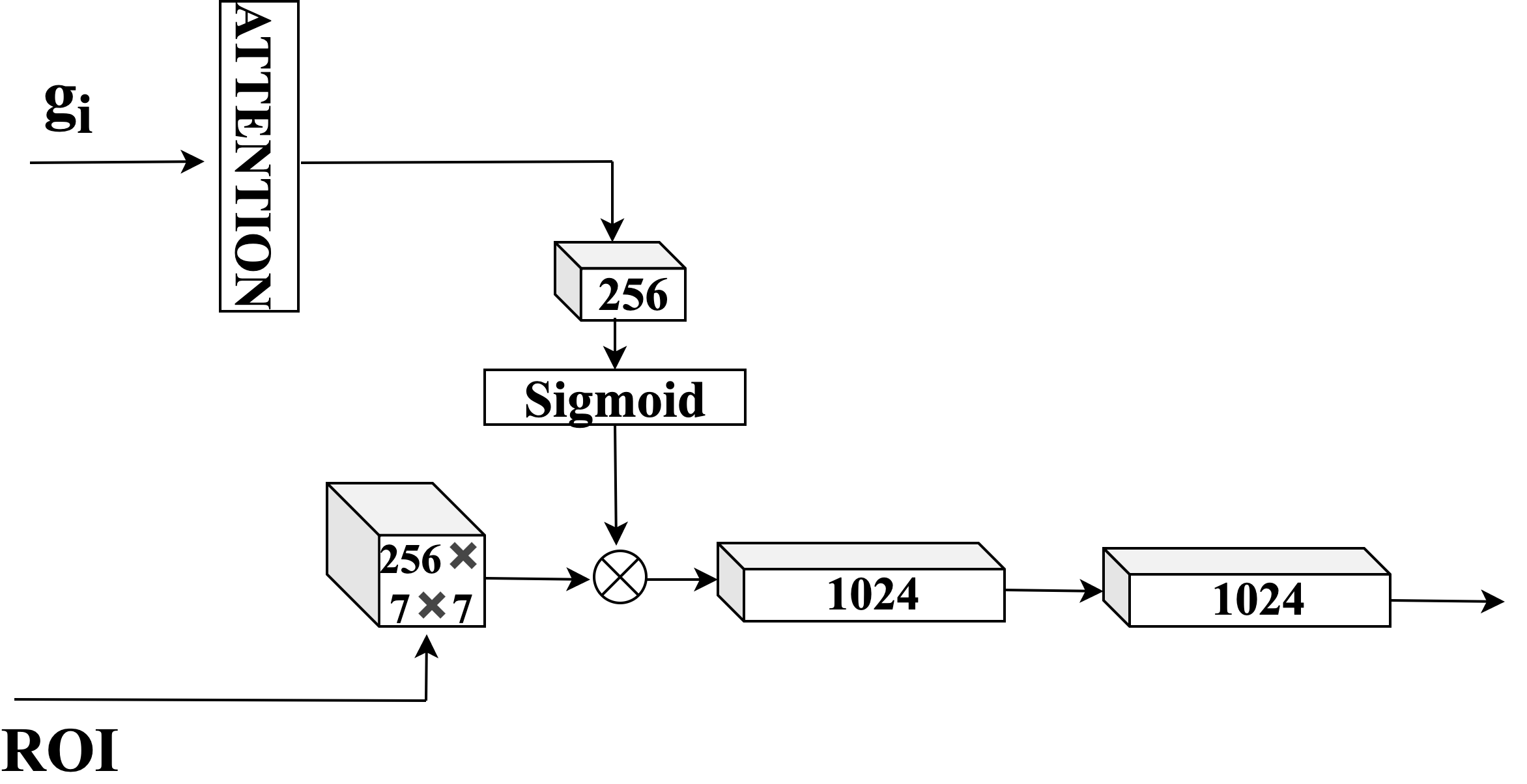}}  \quad \quad \quad \quad
  \subfigure[]{\includegraphics[width=6.3cm,height=3.7cm]{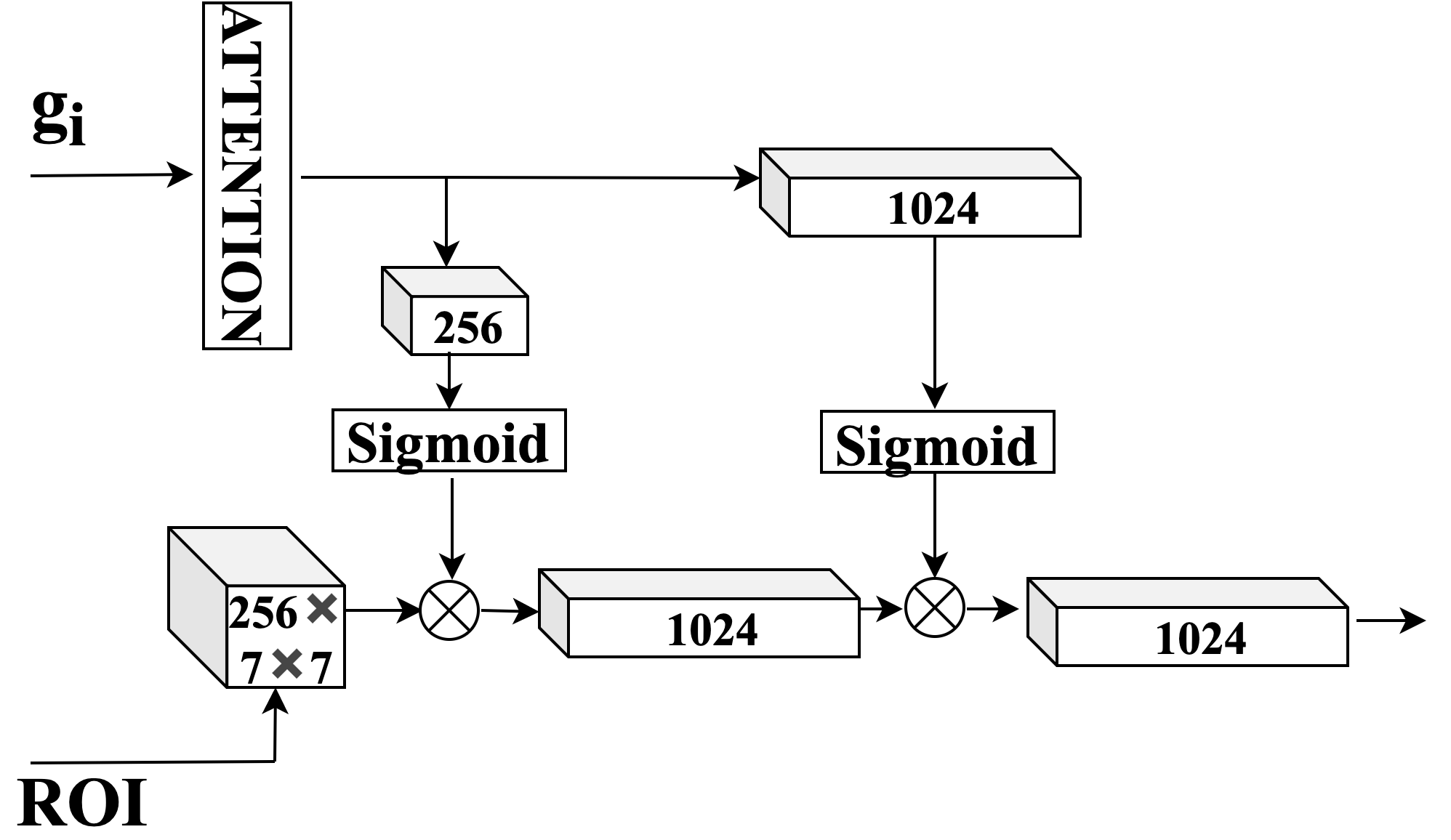}} \\
    \subfigure[]{\includegraphics[width=7cm,height=3.7cm]{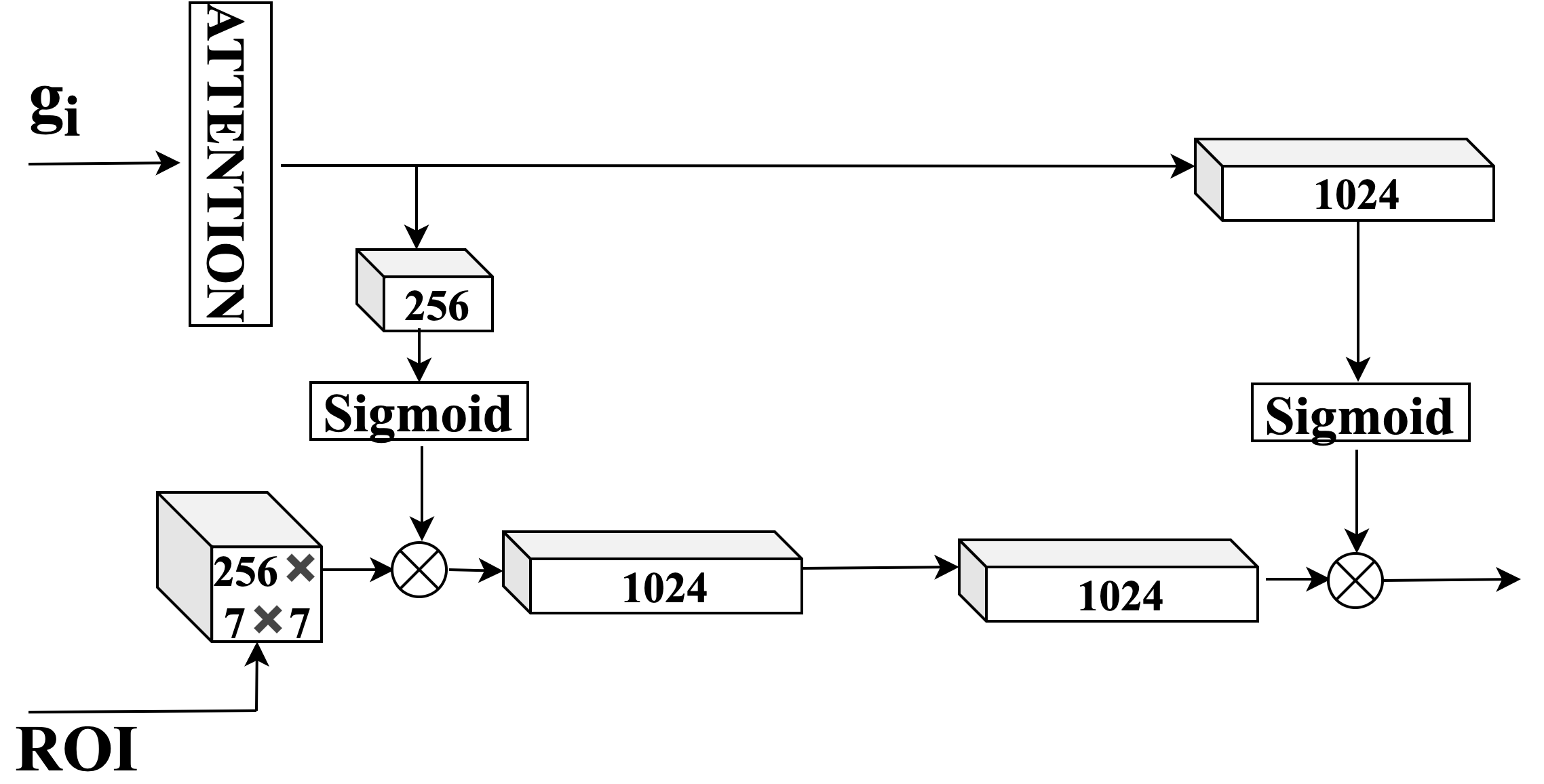}} \quad \quad \quad \quad
  \subfigure[]{\includegraphics[width=6.8cm,height=3.7cm]{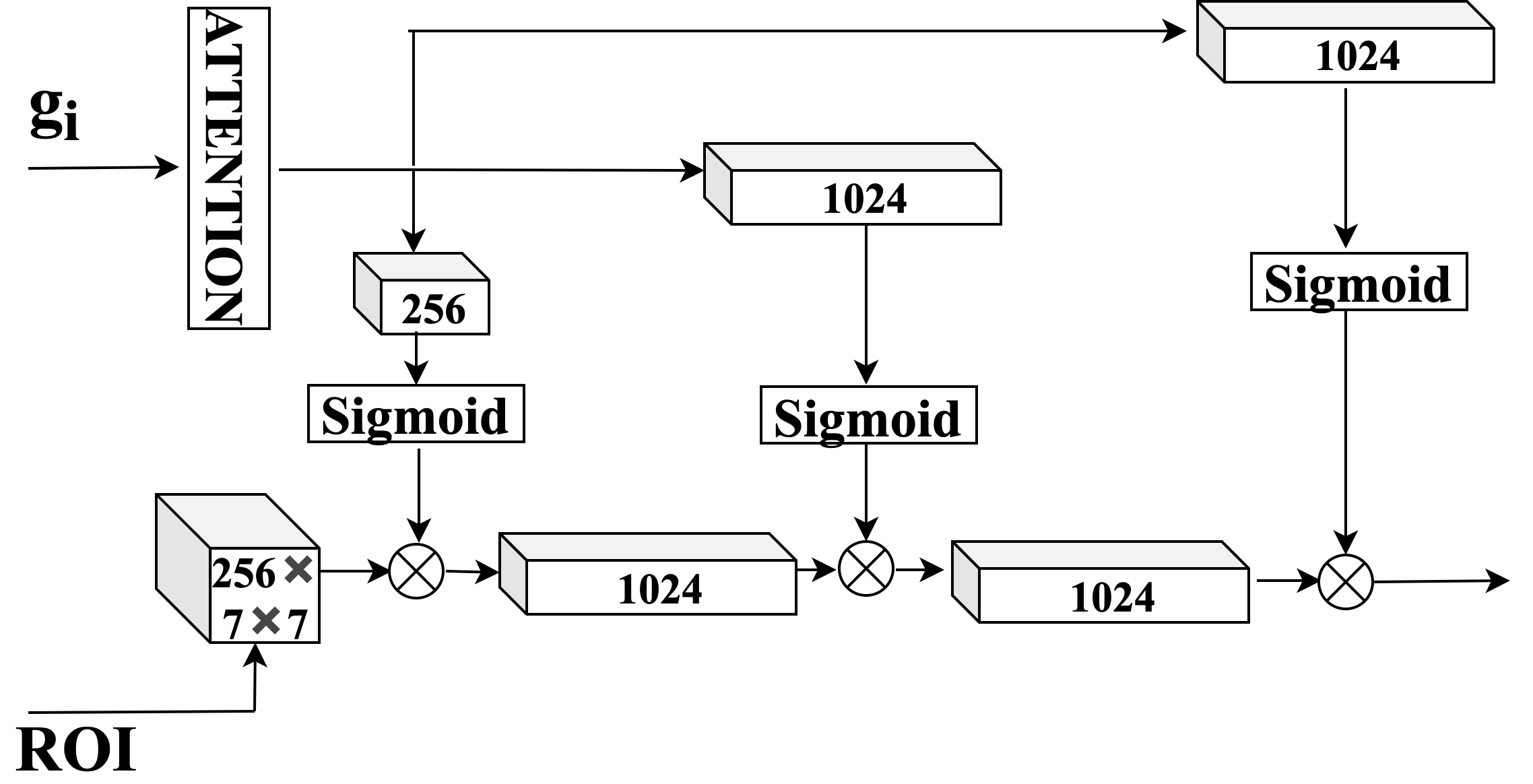}} \\
  \caption{Different attention design choices for context aware. In this Figure, the "ATTENTION" module refers to the composite function of three consecutive operations: global avgpooling and a fully connected layer with ReLU activation function. And $\otimes$ represents the channel-wise multiplication.}
  \label{fig:fig5}
\end{figure*}

{\bf The choices of the attention module: }
For the sake of better integrating global context information and local receptive field information, we explore different attention methods, the results of comparison are shown in Table~\ref{tab:my-table3}. specifically, as shown in Figs.~\ref{fig:fig5}(a), we directly do the channel-wise multiplication between the 256-dimensional global context information output by the squeeze-excitation module and the $256\times7\times7$ feature tensor of the object proposal, and we denote this method as attention on \emph{“conv”}. Further more, on the basis of attention on \emph{“conv”}, we add a new branch to the squeeze-excitation module and increase the output dimension to 1024, then we try two different structures, one is do the channel-wise multiplication between it and the 1024-dimensional output tensor of the first fully connected layer (attention on \emph{"conv+fc1"}, as shown in Figs.~\ref{fig:fig5}(b)) and the other is between it and the 1024-dimensional output tensor of the second fully connected layer (attention on \emph{"conv+fc2"}, as shown in Figs.~\ref{fig:fig5}(c)). In addition, as shown in Figs.~\ref{fig:fig5}(d), we connected three parallel output branches after the squeeze-excitation module, denote as \emph{"conv+fc1+fc2"}. Simultaneously, we also explored different combinations of attention on single \emph{“fc”} layer. The results reported in Table~\ref{tab:my-table3} indicate that assembling global context information multiple times in the box head network will cause ambiguity, and mapping it to the low-dimensional features of the local information can better learn the relationship between the global context and the local receptive field. By using attention on \emph{“conv”} we got the best AP value, which exceeded FPN baseline by 1.5\% on COCO's standard AP metric.

\begin{table}[]
\centering
\caption{Effect of different choices of the attention module on COCO val2017 (\%)}
\label{tab:my-table3}
\setlength{\tabcolsep}{4mm}{
\begin{tabular}{cccc}
\hline
\multicolumn{1}{c}{\bf Attention}  &\multicolumn{1}{c}{\bf AP} &\multicolumn{1}{c}{\bf AP$_{0.5}$} &\multicolumn{1}{c}{\bf AP$_{0.75}$} 
\\ \hline 
\emph{conv}         &  \textbf{38.2} &  \textbf{59.9}  &  \textbf{41.0}   \\
\emph{fc1}          & 37.5    & 58.8     & 40.6      \\
\emph{fc2}          & 37.6 & 59.4  & 40.1   \\
\emph{conv+fc1}     & 37.7 & 59.5  & 40.3   \\
\emph{conv+fc2}     & 37.7 & 59.2  & 40.6   \\
\emph{conv+fc1+fc2} & 37.9    & 59.4     & 40.9      \\
\hline
\end{tabular}}

\end{table}

{\bf Attention with different reduction ratio: }
The comparison in Table~\ref{tab:my-table4} shows the effect of different reduction ratios in the squeeze-excitation module on the performance of our model. Different reduction ratios allow us to explore different capacity and computational cost of the attention module in the network. We can achieve the best results when r is equal to 8, therefore, we use this value in all of our other experiments. 

\begin{table}[]
\centering
\caption{Effect of different reduction ratios r on COCO val2017 (\%)}
\label{tab:my-table4}
\setlength{\tabcolsep}{4mm}{
\begin{tabular}{cccc}
\hline
\multicolumn{1}{c}{\bf ratio}  &\multicolumn{1}{c}{\bf AP} &\multicolumn{1}{c}{\bf AP$_{0.5}$} &\multicolumn{1}{c}{\bf AP$_{0.75}$} 
\\ \hline 
4   & 37.8 & 59.7  & 40.5   \\
8   &  \textbf{38.2}    &  \textbf{59.9}     &  \textbf{41.0}      \\
16  & 37.9    & 59.6     & 40.6      \\ 
\hline
\end{tabular}}

\end{table}


{\bf Bottom-up or Top-down: }
In the previous experiment, we obtained global context information by continuously down-sampling from top to bottom. In this experiment, we replace all newly added $3\times3$ convolutional layers with deconvolutional layers to obtain global context information in a bottom-up manner, and keep other network structures unchanged, the results are shown in Table~\ref{tab:my-table5}.

\begin{table}[]
\centering
\caption{Effect of Bottom-up and Top-down on COCO val2017 (\%)}
\label{tab:my-table5}
\setlength{\tabcolsep}{4mm}{
\begin{tabular}{cccc}
\hline
\multicolumn{1}{c}{\bf Method}  &\multicolumn{1}{c}{\bf AP} &\multicolumn{1}{c}{\bf AP$_{0.5}$} &\multicolumn{1}{c}{\bf AP$_{0.75}$}  
\\ \hline 

Bottom-up & 37.4  & 58.7     & 40.2      \\
Top-down   &  \textbf{38.2}  &  \textbf{59.9}     &  \textbf{41.0}      \\ 
\hline
\end{tabular}}

\end{table}

{\bf Feature recalibration on rpn head: }
In rpn head, in order to determine whether the anchor box generated on the output feature map at different stages of the backbone network contains objects and obtain the corresponding regression coefficients, it is necessary to use the feature map as input for classification and regression tasks respectively. For the purpose of enhancing the global feature information interaction between the input feature map channels, we first use global average pooling to reduce the spatial size of the input feature map, then use a fully connected layer with the same input and output channels for feature refining, and finally it is used for channel-wise multiplication with the input feature map to enhance the feature representation ability, and the results are shown in Table~\ref{tab:my-table7}.

{\bf Additional datasets:}
Next, we investigate whether the benefits of global context information generalise to datasets beyond COCO. For this purpose, we perform experiments with our method on Cityscapes dataset which comprise a collection of 2975 training, 500 validate and 1525 test $2048\times1024$ pixel RGB images, and labelled with 8 classes. We train our model a total of 64 epochs, the learning rate is initially set to 0.0025 and drops by a factor of 10 after 48 epochs. We set the initial pooling size to (128, 256). The shorter edges of the images are randomly sampled from [800, 1024] for reducing overfitting, other parameters are the same as those set in the experiment on COCO dataset. From Table~\ref{tab:my-table6} we observe that our method achieves a better AP value (1.2\% improvement) than FPN baseline on Cityscapes datasets, which further illustrates the robustness of our method.

\begin{table}[!h]
\centering
\caption{Comparisons with FPN baseline on Cityscapes datasets with ResNet-50 backbone (\%)}
\label{tab:my-table6}
\setlength{\tabcolsep}{4mm}{
\begin{tabular}{ccccccc}
\hline
\multicolumn{1}{c}{\bf Method}   &\multicolumn{1}{c}{\bf AP} &\multicolumn{1}{c}{\bf AP$_{0.5}$} &\multicolumn{1}{c}{\bf AP$_{0.75}$} &\multicolumn{1}{c}{\bf AP$_{s}$} &\multicolumn{1}{c}{\bf AP$_{m}$} &\multicolumn{1}{c}{\bf AP$_{l}$}
\\ \hline 

FPN baseline & 36.2  & 63.6     & 34.8  &10.6 &30.9 &51.3    \\
GCA (ours)   &  \textbf{37.4}  &  \textbf{63.8}     &  \textbf{38.2}  & \textbf{13.0} & \textbf{31.9} & \textbf{52.2}    \\ 
\hline
\end{tabular}}

\end{table}

\subsection{Main Results}

{\bf Comparison with FPN Baselines on COCO val2017:}
In Table~\ref{tab:my-table7}, we compare the performance of our method with FPN baselines on COCO val2017, where (+rpn head) means to deploy feature calibration on rpn head. From Table~\ref{tab:my-table7} we can see that our method can achieve continuous gains on different backbone networks (1.5\% improvement with ResNet-50 and 0.9\% improvement with ResNet-101). Compared with FPN baselines, our method is better at detecting small and medium targets, which is due to the increased connection between global context and local information. Furthermore, by recalibrating the input features of rpn head to strengthen the relationship between global features, our model's detection ability for larger objects has also been greatly improved. Meanwhile, from Table~\ref{tab:my-table7} we can see that the effect of global feature recalibration is not obvious in rpn head(0.1\% improvement with ResNet-50 and 0.3\% improvement with ResNet-101), since the rpn head uses the entire feature map of the backbone network output as input. Therefore, in rpn head, not that the global context information is lost, but that the information exchange between channels is lacking. However, in roi head, which using the local feature map after roi align cropping as input will lose a lot of background information. Consequently, using our proposed CGA to compensate for the missing global context information in roi head can make the baseline model obtain a larger performance improvement, which also experimentally validates the rationality of our hypothesis and the effectiveness of our proposed method.

{\bf Comparison with State-of-the-art Methods on COCO test-dev:} 
Table~\ref{tab:my-table8} shows the comparison between our method with the state-of-the-art methods on MS COCO test-dev2017. In Table~\ref{tab:my-table8}, we also compare the performance of our method with FPN baselines and Double-Head~\citep{wu2020rethinking} 
on COCO test-dev, where Double-Head (GCA) means to assemble our method on Double-Head RCNN~\citep{wu2020rethinking} based on mmdetection~\citep{mmdetection}, note that in this method we only applied our GCA to the fully connected head and we re-implemented Double-Head RCNN using two Titan xp GPUs with one image per GPU (schedule\_1x). Our method can achieve continuous gains on different backbone networks (0.9\% improvement with ResNet-101) and model (0.5\% compared to Double-Head). 
\begin{table*}[!h]
\begin{center}
\caption{Object detection results (bounding box AP) on COCO val2017(\%).}
\label{tab:my-table7}
\setlength{\tabcolsep}{3mm}{
\begin{tabular}{cccccccc}
\hline
\multicolumn{1}{c}{\bf Method}  &\multicolumn{1}{c}{\bf Backbone} &\multicolumn{1}{c}{\bf AP} &\multicolumn{1}{c}{\bf AP$_{0.5}$} &\multicolumn{1}{c}{\bf AP$_{0.75}$} &\multicolumn{1}{c}{\bf AP$_{s}$} &\multicolumn{1}{c}{\bf AP$_{m}$} &\multicolumn{1}{c}{\bf AP$_{l}$}
\\ \hline 
FPN baseline~\citep{lin2017feature} & ResNet-50     & 36.8 & 58.0  & 40.0   & 21.2 & 40.1 & 48.8 \\
GCA (ours)   & ResNet-50     & 38.2 & 59.9  & 41.0   & 22.7 &  \textbf{42.0} & 49.0 \\
GCA (ours+rpn head)   & ResNet-50      &  \textbf{38.3} &  \textbf{60.1}  &  \textbf{41.0}   &  \textbf{22.9} & 41.8 &  \textbf{49.2} \\
\hline 
FPN baseline~\citep{lin2017feature} & ResNet-101    & 39.1 & 61.0  & 42.4   & 22.2 & 42.5 & 51.0 \\
GCA (ours)   & ResNet-101    & 39.7 & 61.0  & 43.3   & 23.0 & 43.7   & 51.3    \\ 
GCA (ours+rpn head)   & ResNet-101    &  \textbf{40.0} &  \textbf{61.5}  &  \textbf{43.4}   &  \textbf{23.5} &  \textbf{43.9}   &  \textbf{52.4}    \\ 
\hline
\end{tabular}}
\end{center}

\end{table*}

\begin{table*}[!h]
\begin{center}
\caption{Object detection results (bounding box AP) on COCO test-dev(\%).}
\label{tab:my-table8}
\setlength{\tabcolsep}{3mm}{
\begin{tabular}{cccccccc}
\hline
\multicolumn{1}{c}{\bf Method}  &\multicolumn{1}{c}{\bf Backbone} &\multicolumn{1}{c}{\bf AP} &\multicolumn{1}{c}{\bf AP$_{0.5}$} &\multicolumn{1}{c}{\bf AP$_{0.75}$} &\multicolumn{1}{c}{\bf AP$_{s}$} &\multicolumn{1}{c}{\bf AP$_{m}$} &\multicolumn{1}{c}{\bf AP$_{l}$}
\\ \hline 
Deep Regionlets ~\citep{xu2018deep} & ResNet-101  & 39.3 & 59.8  & -   & 21.7 & 43.7 & 50.9 \\
Mask R-CNN ~\citep{He_2017}  & ResNet-101  & 39.8 & 62.3  & 43.4   & 22.1 & 43.2 & 51.2 \\
IOU-Net ~\citep{jiang2018acquisition} & ResNet-101  & 40.6 & 59.0  & -   & - & - & - \\
Soft-NMS ~\citep{bodla2017soft} & Aligned-Inception-ResNet  & 40.9 & 62.8  & -   & 23.3 & 43.6 & 53.3 \\
LTR ~\citep{tan2019learning} & ResNet-101  & 41.0 & 60.8  & 44.5   & 23.2 & 44.5 & 52.5 \\
Fitness NMS ~\citep{tychsen2018improving} & ResNet-101  & 41.8 & 60.9  & 44.9   & 21.5 & 45.0 &  \textbf{57.5} \\
\hline
FPN baseline~\citep{lin2017feature} & ResNet-101  & 39.1 & 60.5  & 42.4   & 22.0 & 42.2 & 49.3 \\
GCA (ours)   & ResNet-101    & 40.0 & 61.6  & 43.5   & 22.8 & 43.2   & 50.3    \\
Double-Head~\citep{wu2020rethinking}   & ResNet-101     & 41.6 & 62.0  & 45.7   & 23.8 & 44.8 & 52.7 \\
Double-Head (GCA)   & ResNet-101    &  \textbf{42.1} &  \textbf{63.0}  &  \textbf{45.9}   &  \textbf{24.4} &  \textbf{45.2}   &  53.2    \\
\hline
\end{tabular}}
\end{center}

\end{table*}
{\bf Lightweight version:}
In order to reduce the amount of parameters and model complexity to obtain better precision and speed trade-off, in this experiment, we present a lightweight version of our method. Specifically, for feature recalibration on rpn head, we simply use a $1\times1$ convolutional layer to replace the newly added fully connected layer, and for the global information extraction of the roi head , we directly use the feature maps at different stages in the feature pyramid of $\displaystyle \{p2, p3, p4, p5\}$ to obtain their corresponding spatial statistics through global average pooling, and then fuse them by element-wise sum, and after that only one context aware module is retained to refine the fused global features, finally, we do the channel-wise multiplication between the global context descriptor and the input roi feature. As can be seen from the Table~\ref{tab:my-table9}, compared to FPN baseline, our lightweight version of GCA can achieve good performance improvements(+0.8\%) while a similar inference speed is maintained.
\begin{table*}[!h]
\begin{center}
\caption{Comparisons with lightweight version on COCO val2017(Runtime is measured on a single NVIDIA titan xp GPU).}
\label{tab:my-table9}
\setlength{\tabcolsep}{3mm}{
\begin{tabular}{cccccc}
\hline
\multicolumn{1}{c}{\bf Method}  &\multicolumn{1}{c}{\bf Backbone} &\multicolumn{1}{c}{\bf AP} &\multicolumn{1}{c}{\bf AP$_{0.5}$} &\multicolumn{1}{c}{\bf AP$_{0.75}$}  &\multicolumn{1}{c}{\bf Runtime FPS}
\\ \hline
FPN baseline~\citep{lin2017feature} & ResNet-50  & 36.8 & 58.0  & 40.0  & 16.3 \\
GCA (ours)   & ResNet-50    & 38.3 & 60.1  & 41.0    & 13.1    \\
GCA (lightweight version)   & ResNet-50    & 37.6 & 59.3  & 40.7   & 16.1    \\
\hline
\end{tabular}}
\end{center}

\end{table*}

\section{Conclusion}
\label{conclusion}

In this paper, we propose Global Context Aware (GCA) RCNN to learn the potential relationship between image background and foreground by integrating global context information with local receptive field information of roi, and different from the attention mechanism on the convolution level for feature recalibration, we extend it to the network pipeline to strengthen the connection between local and global information. Experiments on the COCO and Cityscapes datasets have verified the effectiveness of our method, and we also hope that our method will be helpful to other scholars.


\bibliographystyle{elsarticle-num} 
\bibliography{iclr2021_conference}





\end{document}